\documentclass[letterpaper, 10 pt, conference]{ieeeconf}  
\usepackage{microtype}
\usepackage{graphicx}
\usepackage{subcaption}
\usepackage{booktabs} 
\usepackage{amsmath, amssymb, amsfonts}
\usepackage{graphicx}
\usepackage{booktabs}
\usepackage{hyperref}
\usepackage{array}
\usepackage{cite}
\usepackage{color}
\usepackage{multirow}
\usepackage{tabularx}

\IEEEoverridecommandlockouts                              

\title{\LARGE \bf
Interaction-Aware Whole-Body Control for Compliant Object Transport
}

\author{Hao Zhang$^{1,2}$, Yves Tseng$^{1}$, Ding Zhao$^{2}$ and H. Eric Tseng$^{1}$
\thanks{*This work was supported by the University of Texas at Arlington and Carnegie Mellon University. Correspondance to H. Eric Tseng (hongtei.tseng@uta.edu) and Ding Zhao (dingzhao@cmu.edu).}
\thanks{$^{1}$Hao Zhang, Yves Tseng and H. Eric Tseng are with Department of Electrical Engineering, the University of Texas at Arlington, 76010 Arlington, USA
        {\tt\small haoz4@andrew.cmu.edu; ytseng@duck.com; hongtei.tseng@uta.edu}}%
\thanks{$^{2}$Hao Zhang and Ding Zhao are with Department of Mechanical Engineering, Carnegie Mellon University, 15213 Pittsburgh, USA
        {\tt\small haoz4@andrew.cmu.edu; dingzhao@cmu.edu}}%
}

\begin{document}

\maketitle
\thispagestyle{empty}
\pagestyle{empty}

\begin{abstract}
Cooperative object transport in unstructured environments remains challenging for assistive humanoids because strong, time-varying interaction forces can make tracking-centric whole-body control unreliable, especially in close-contact support tasks. 
This paper proposes a bio-inspired, interaction-oriented whole-body control (IO-WBC) that functions as an artificial cerebellum - an adaptive motor agent that translates upstream (skill-level) commands into stable, physically consistent whole-body behavior under contact. This work structurally separates upper-body interaction execution from lower-body support control, enabling the robot to maintain balance while shaping force exchange in a tightly coupled robot--object system. 
A trajectory-optimized reference generator provides a kinematic prior, while a reinforcement learning (RL) policy governs body responses under heavy-load interactions and disturbances. 
The policy is trained in simulation with randomized payload mass/inertia and external perturbations, and deployed via asymmetric teacher--student distillation so that the student relies only on proprioceptive histories at runtime. Extensive experiments demonstrate that IO-WBC maintains stable whole-body behavior and physical interaction even when precise velocity tracking becomes infeasible, enabling compliant object transport across a wide range of scenarios.
\end{abstract}


\section{Introduction}

The deployment of humanoid robots in unstructured, human-centric environments necessitates a robust and versatile whole-body control (WBC) framework that functions as a cerebellum. For complex tasks such as human--robot collaboration (HRC) \cite{khatib1999robots}, the primary challenge lies in managing the bidirectional physical interaction between the robot and its partner \cite{fang2023human}. This cerebellum must reliably translate high-level intent generated by the brain into stable whole-body behavior under continuous physical interaction \cite{wang2025navigating}, requiring a high degree of active compliance to safely absorb human-induced disturbances while maintaining the necessary stiffness for task execution \cite{parietti2016design}. In particular, collaborative object transport requires the robot to apply sustained forces to an external payload while simultaneously maintaining balance \cite{solanes2018human}, resulting in a tightly coupled robot--object dynamic system \cite{an2023multi}. Under such heavy-load conditions, effective control is determined not solely by tracking accuracy \cite{wang2025human}, but by the ability to remain stable, compliant, and responsive in the presence of strong, non-conservative interaction forces \cite{khoramshahi2020dynamical}.

\begin{figure}[t]
    \centering
    \includegraphics[width=\linewidth]{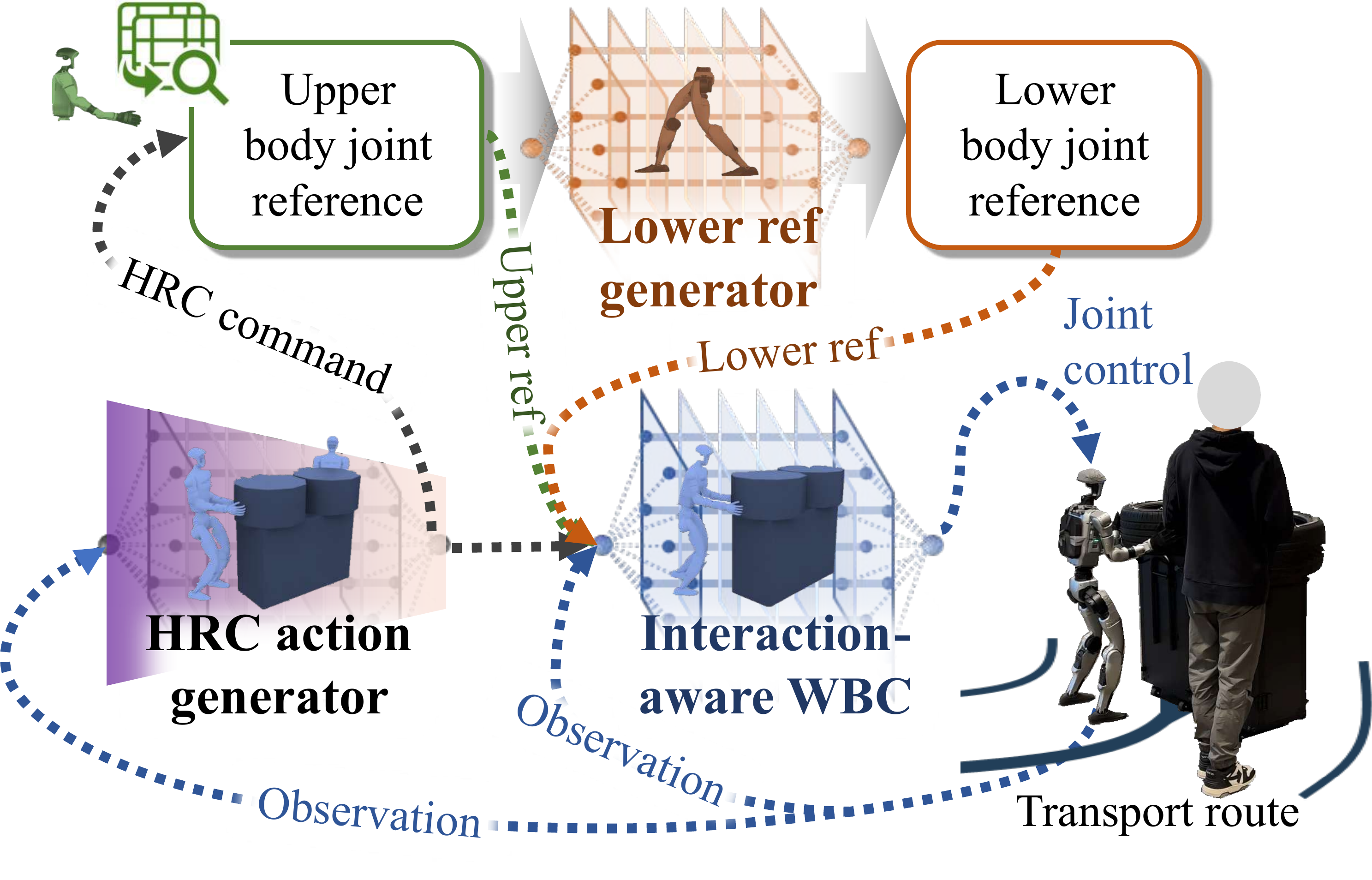}
    \caption{The proposed IO-WBC architecture that bridges skill-level HRC commands with low-level execution through a kinematic-prior-based RG, enabling stable coordination between upper-body interaction and lower-body support under strong robot--object coupling.}
    \label{fig:system_overview}
\end{figure}

A competent WBC must therefore coordinate flexible upper-body interaction with lower-body dynamic support \cite{sentis2007synthesis}. By regulating force exchange and body momentum, the WBC provides high-level HRC planners with the control authority necessary to modulate center-of-mass (CoM) motion and interaction forces. This capability is particularly critical in collaborative transport, where non-conservative force exchange with a human partner or a heavy object can destabilize traditional control loops that rely on weak coupling assumptions \cite{buerger2007complementary}. Recent advances in deep reinforcement learning (DRL) have demonstrated strong potential for whole-body control in high-dimensional, nonlinear systems \cite{peng2018deepmimic}. DRL-based approaches enable agile motor skill acquisition \cite{peng2021amp}, yet extending these capabilities to physical human--robot collaboration introduces additional challenges. Humans naturally decouple upper-body interaction from lower-body support while preserving a tight functional coupling between them, allowing interaction forces to be absorbed and redirected through the whole body \cite{luo2024enhancing}. Inspired by this biological principle, a hierarchical WBC architecture can prevent conflicts between interaction objectives and balance maintenance, enabling stable behavior even when external forces dominate the system dynamics \cite{ait2026human}.

Despite substantial progress in humanoid motion control, existing frameworks such as ALBERT \cite{bellicoso2019albert}, AMO \cite{li2025amo}, and optimization-based methods \cite{ames2019control} primarily emphasize accurate posture or trajectory tracking under relatively weak interaction assumptions. Although recent works have incorporated learning-based components to improve robustness \cite{boffi2021learning}, these approaches often degrade under high-pressure physical interactions, such as collaborative transport of heavy payloads with human partners \cite{haight2025harl}. In such scenarios, the object mass and contact forces induce significant disturbances, rendering precise velocity tracking physically unfeasible \cite{manek2019learning}. Enforcing strict tracking objectives in these regimes can lead to unstable or overly aggressive behaviors, highlighting a fundamental limitation of robot-centric WBC formulations.

Motivated by this, we investigate a WBC paradigm that prioritizes stable force application and coupled motion consistency over exact tracking accuracy when operating under heavy-load interactions. We introduce the interaction-oriented whole-body control (IO-WBC) in a hierarchical framework, which explicitly accounts for robot--object coupling and supports serial execution between skill-level HRC planning and low-level physical interaction. The main contributions are summarized as follows: 1) a hierarchical whole-body control architecture is proposed, composed of a trajectory-optimized reference generator (RG) and an IO-WBC policy. The RG provides kinematic priors from abstract HRC commands, while the IO-WBC policy regulates whole-body stability and force application under strong interaction coupling; 2) we develop a physics-aware training environment that models heavy payload dynamics, external perturbations, and contact uncertainties. Within this environment, an asymmetric teacher--student distillation scheme enables the policy to infer interaction dynamics without relying on privileged sensing at deployment; 3) comprehensive experiments demonstrate that IO-WBC effectively serves as a cerebellum within a multi-agent reinforcement learning (MARL) framework, enabling stable whole-body force application and cooperative transport across diverse tasks and payload conditions.

\section{Related Work}

\subsection{Whole-Body Control and Reinforcement Learning}
Traditional WBC for humanoid robots has long relied on inverse kinematics or quadratic programming (QP) to manage multi-contact stability and task-space objectives \cite{dietrich2015whole}. While these optimization-centric methods provide theoretical guarantees, they often struggle with the non-linear, high-frequency disturbances inherent in heavy-load physical interactions \cite{koppula2015anticipating}. Recent advancements in DRL have demonstrated superior adaptability in learning agile motor skills \cite{hwangbo2019learning}. Notably, frameworks such as AMO \cite{li2025amo} leverage adaptive optimization to enhance humanoid dexterity. However, these methods primarily focus on single-agent performance and lack an explicit mechanism to decouple upper-body interaction from lower-body support, which is essential for sustaining effective force application during collaborative tasks.

\subsection{Bionic-Inspired Hierarchical Body Control}
Physical collaboration in transport tasks has traditionally been modeled via leader-follower paradigms, frequently necessitating explicit force-torque sensing \cite{lawitzkystrategies}. To achieve more practical and robust human-robot cooperation, researchers have investigated bionic-inspired control structures that mimic the functional separation of the brain and cerebellum. Some approaches utilize stability certificates \cite{chow2018lyapunov} to provide safety guarantees for dynamic systems \cite{khalil2002nonlinear}. Nevertheless, the effective serial execution of task-level intent (brain) and interaction-aware stabilization (cerebellum) remains under-explored for heavy-load scenarios \cite{ren2024whole}. Addressing this, our work introduces a hierarchical structure that decouples upper-body interaction from lower-body support to enhance postural resilience under significant external payloads.

\subsection{Motion Synthesis and Kinematic Priors}
A primary challenge in DRL-based humanoid control is the out-of-distribution (OOD) vulnerability when robots encounter unforeseen interaction forces \cite{sikchi2021lyapunov}. Standard motion synthesis frequently fails to account for the specific hardware constraints and contact-rich states required for collaborative transport. Previous research has explored hierarchical motion priors \cite{peng2021amp, li2025amo} and offline pre-training methodologies \cite{pmlr-v162-wang22ai} to map abstract high-level commands into a kinematically feasible subspace. Although such priors provide a structured starting point for policy exploration, existing frameworks often do not link these kinematic references with interaction-aware execution, leading to distribution shifts when learning complex, coupled collaborative behaviors.

\section{Methodology}
\label{sec:method}

\begin{figure*}[t]
    \centering
    \includegraphics[width=0.93\textwidth]{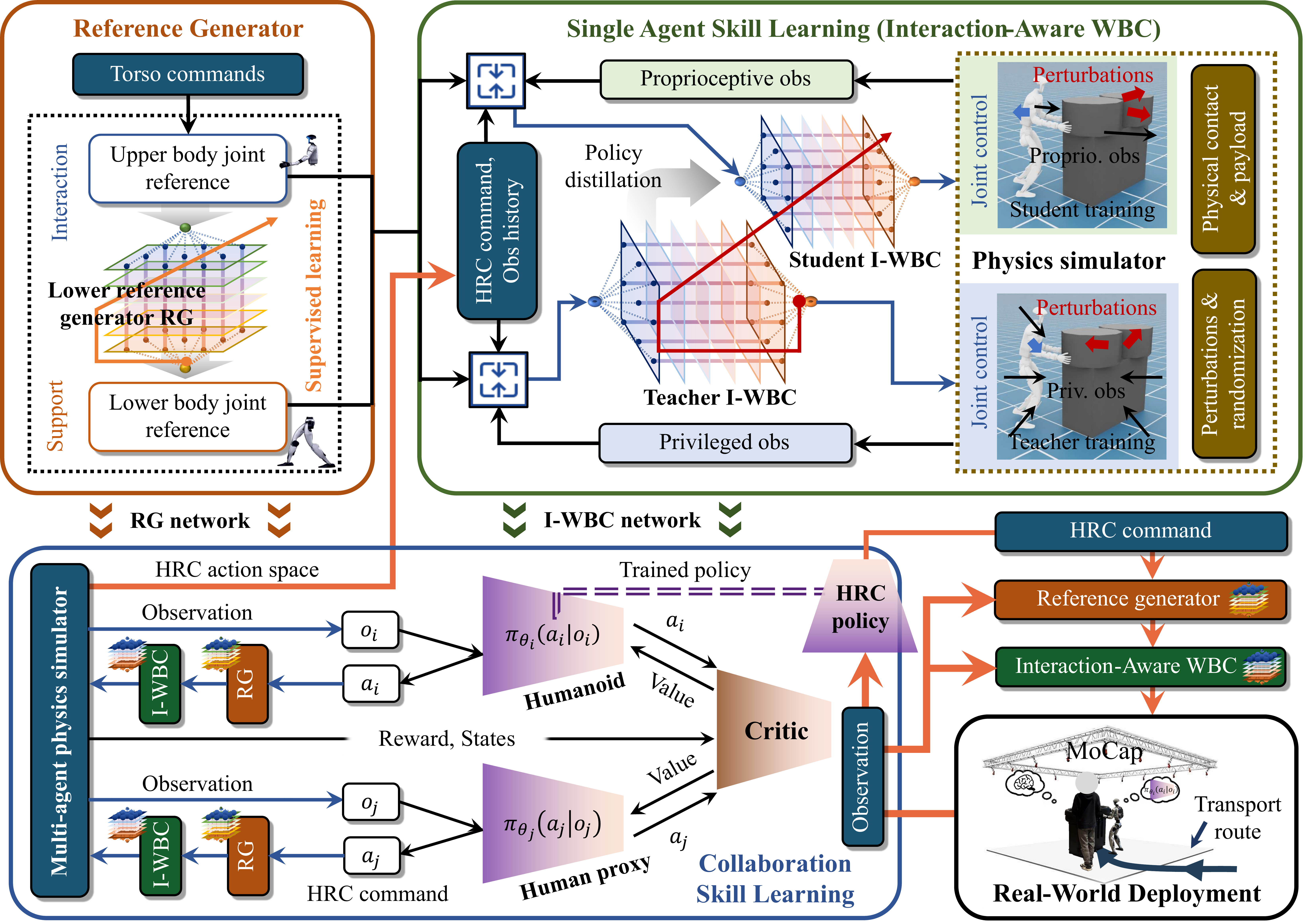}
    \caption{The comprehensive learning and execution pipeline of IO-WBC. The RG is trained via supervised learning to provide kinematic priors. The interaction-oriented WBC policy is developed through teacher-student distillation, where a privileged teacher guides a student policy to decode interaction dynamics from proprioceptive history.}
    \label{fig:algorithm_framework}
\end{figure*}

We propose the interaction-oriented whole-body control framework, as detailed in Fig. \ref{fig:algorithm_framework}, designed to address the challenges of motion coordination and postural stability for humanoid robots during dynamic collaborative carrying tasks. The framework utilizes a cascaded architecture that transforms tactical decisions at skill-level into low-level driving torques through kinematic priors and local task-space resolution.

\subsection{System Modeling and Cascaded Control Architecture}

We model the humanoid robot as a branched dynamic system. The generalized coordinate vector is defined as:
\begin{equation}
    \mathbf{q} = [ \mathbf{x}_{b}^{T}, \mathbf{R}_{b}^{T}, \boldsymbol{\theta}_{\mathcal{I}}^{T}, \boldsymbol{\theta}_{\mathcal{S}}^{T} ]^{T}
\end{equation}
where $\mathbf{x}_{b}$ and $\mathbf{R}_{b} \in SO(3)$ denote the position and rotation matrix of the floating base in the world frame, respectively. Besides, $\boldsymbol{\theta}_{\mathcal{I}} \in \mathbb{R}^{n_i}$ represents the joint sequence of the upper-body interaction branch, and $\boldsymbol{\theta}_{\mathcal{S}} \in \mathbb{R}^{n_s}$ represents the joint sequence of the lower-body and waist support branch responsible for balance and height regulation. Note that the separation between interaction branch $\mathcal{I}$ and support branch $\mathcal{S}$ is architectural rather than dynamically block-diagonal. 
Coupling still exists through the full-body dynamics, but the control responsibilities are distributed across branches to handle non-conservative interaction forces. The task execution process is driven by the three-tier cascaded structure:

1) Skill policy layer (HRC): composed of an MARL policy, this layer outputs an 11-dimensional command set $\mathcal{C} = \{ \mathbf{v}_{coll}, \mathbf{\sigma}_{base}, \Delta \mathbf{p} \}$. Here, $\mathbf{v}_{coll} = [v_x, v_y, \omega_{yaw}]^T$ is the collaborative system movement velocity and yaw; $\mathbf{\sigma}_{base} = [H_{CoM}, \alpha_{pitch}]^T$ represents the target center-of-mass (CoM) height and torso pitch angle; and $\Delta \mathbf{p} \in \mathbb{R}^6$ denotes the 3D spatial offsets of the bilateral wrists relative to the base. This layer abstracts the multi-agent task coordination into reachable task-space objectives.

2) Reference generation layer (RG): this layer receives commands from the HRC layer and resolves the kinematic reference posture for the support branch: $\boldsymbol{\theta}_{\mathcal{S}}^{ref} = \phi_{RG}(\boldsymbol{\theta}_{\mathcal{I}}^{ref}, \mathbf{\sigma}_{base})$. Through a pre-trained mapping, this module ensures that the lower body possesses predictive adjustment capabilities in a kinematic sense. The RG functions as a feasibility prior, pre-shaping the support branch to ensure the CoM projection remains within the geometric stability boundary, thereby reducing the search space for the subsequent reinforcement learning policy.

3) Interaction execution layer (IO-WBC): at the core of this layer is an interaction-aware RL policy $\pi_{low}$, which maps high-level commands $\mathcal{C}$, the kinematic prior $\boldsymbol{\theta}_{\mathcal{S}}^{ref}$, and proprioceptive history $\mathbf{H}_t$ to optimal joint-space adaptations $\mathbf{a}_t$. Unlike standard tracking controllers, this RL policy is trained to implicitly infer interaction dynamics, generating real-time residual corrections that compensate for heavy-load perturbations and non-conservative contact forces. These learned increments $\mathbf{a}_t$ are then superposed onto the RG prior and converted into driving torques $\mathbf{\tau}$ via low-level PD controllers: $\mathbf{\tau}_{\mathcal{S}} = K_p (\boldsymbol{\theta}_{\mathcal{S}}^{ref} + \mathbf{a}_t - \boldsymbol{\theta}_{\mathcal{S}}) - K_d \dot{\boldsymbol{\theta}}_{\mathcal{S}}$. By leveraging RL for reactive execution, the IO-WBC framework can maintain stable whole-body behavior even in regimes where analytical tracking becomes physically infeasible.

\subsection{Kinematic Mapping and Reference Generator}

In the IO-WBC architecture, the generation of the upper-body reference configuration $\boldsymbol{\theta}_{\mathcal{I}}^{ref}$ is based on the wrist spatial offsets $\Delta \mathbf{p}$ provided by the HRC layer. To ensure operational precision, we formulate this as a local inverse kinematics (IK) resolution process embedded within the control loop.

We define the target task-space pose of the bilateral wrists $\mathbf{x}_{wrist} = [ \mathbf{x}_{L}^{T}, \mathbf{x}_{R}^{T} ]^{T}$ as:
\begin{equation}
    \mathbf{x}_{wrist} = \mathbf{x}_{task\_base} + \Delta \mathbf{p}
\end{equation}
where $\mathbf{x}_{task\_base} \in \mathbb{R}^6$ is the pre-defined base grasping pose according to the current carrying task. The corresponding upper-body joint reference $\boldsymbol{\theta}_{\mathcal{I}}^{ref}$ is resolved via a differential IK model:
\begin{equation}
    \Delta \boldsymbol{\theta}_{\mathcal{I}} = \mathbf{J}_{\mathcal{I}}^{\dagger}(\boldsymbol{\theta}_{\mathcal{I}}) \left( \mathbf{x}_{wrist} - \mathcal{FK}_{\mathcal{I}}(\boldsymbol{\theta}_{\mathcal{I}}) \right)
\end{equation}
where $\mathbf{J}_{\mathcal{I}}^{\dagger}$ denotes the Jacobian pseudo-inverse matrix of the interaction branch, and $\mathcal{FK}_{\mathcal{I}}$ is its forward kinematics operator. The resolved $\boldsymbol{\theta}_{\mathcal{I}}^{ref}$ is simultaneously fed into the RG module and the bottom-level policy $\pi_{low}$. This design enables IO-WBC to compensate for dynamic deviations caused by payloads using whole-body redundancy while maintaining end-effector precision.

To ensure the lower body can compensate for CoM shifts caused by upper-body postures during collaboration, we utilize physics-model-driven trajectory optimization to construct the RG pre-training dataset. Inspired by the AMO framework, the RG module encodes the robot's kinematic limits and static stability boundaries through supervised learning on large-scale trajectory data. The problem is formulated as a multi-contact optimal control problem:
\begin{equation}
\begin{aligned}
    \min_{\mathbf{q}, \mathbf{\tau}} & \int_{0}^{T} \left( \| \text{CoM}(\mathbf{q}) - \text{CoM}_{ref} \|^2 + \| \mathbf{\tau} \|^2 + \| \dot{\mathbf{q}} \|^2 \right) dt \\
    \text{s.t.} & \quad \mathbf{M}(\mathbf{q})\ddot{\mathbf{q}} + \mathbf{C}(\mathbf{q}, \dot{\mathbf{q}})\dot{\mathbf{q}} + \mathbf{G}(\mathbf{q}) = \sum \mathbf{J}_{i}^T(\mathbf{q}) \mathbf{f}_{i} + \mathbf{S}^T \mathbf{\tau}
\end{aligned}
\end{equation}
where $\mathbf{M}, \mathbf{C}, \text{and } \mathbf{G}$ represent the system's mass matrix, Coriolis terms, and gravity terms, respectively. The pre-trained RG module is frozen during the real-time training phase, providing a robust lower-body reference benchmark for reinforcement learning.

\subsection{Learning Strategy and Asymmetric Distillation}

The execution of the interaction-aware policy is formulated as a Markov decision process (MDP). To achieve compliant collaboration without external force sensors, we leverage the coupling between object geometric states and proprioceptive feedback to construct a closed-loop control law. The fundamental premise is that external interaction forces $\mathbf{F}_{ext}$ introduce measurable deviations in the robot's joint tracking and base acceleration, which serve as indirect observations of the coupled system dynamics.

1) Observation and action space:
the observation vector for the bottom-level policy at time $t$ is defined as $\mathbf{s}_t \in \mathbb{R}^{n_s}$:
\begin{equation}
    \mathbf{s}_t = [\mathbf{c}_t, \mathbf{\Phi}_{b}, \boldsymbol{\omega}_{b}, \mathbf{q}, \dot{\mathbf{q}}, \phi_t, \mathbf{a}_{t-1}, \boldsymbol{\theta}_{\mathcal{S}}^{ref}, \mathbf{x}_{obj}, \dot{\mathbf{x}}_{obj} ]
\end{equation}
where $\mathbf{c}_t$ is the skill-level HRC command; $\mathbf{\Phi}_{b}$ and $\boldsymbol{\omega}_{b}$ denote the base Euler angles and angular velocities; $\mathbf{q}, \dot{\mathbf{q}}$ are the whole-body joint positions and velocities; $\phi_t$ is the gait phase signal; $\mathbf{a}_{t-1}$ is the previous action; and $\boldsymbol{\theta}_{\mathcal{S}}^{ref}$ is the support reference generated by the RG module. Crucially, the observation includes the object's egocentric geometry $\mathbf{x}_{obj}$ and its first-order derivative $\dot{\mathbf{x}}_{obj}$, enabling the policy to implicitly perceive interaction trends through state deviations. 

To account for the high-inertia effects of heavy payloads, we utilize a proprioceptive history buffer $\mathbf{H}_t = \{ \mathbf{s}_{t-k}, \dots, \mathbf{s}_t \}$. This history captures the time-series response of the actuators, allowing the policy to differentiate between internal joint friction and external resistance. The action space $\mathbf{a}_t \in \mathbb{R}^{15}$ comprises joint position increments for the support branch $\mathcal{S}$, specifically 12 leg joints and 3 waist motors. The final driving torques are regulated via a PD scheme: 
\begin{equation}
    \boldsymbol{\tau}_{\mathcal{S}} = K_p (\boldsymbol{\theta}_{\mathcal{S}}^{ref} + \mathbf{a}_t - \boldsymbol{\theta}_{\mathcal{S}}) - K_d \dot{\boldsymbol{\theta}}_{\mathcal{S}}
\end{equation}

In this formulation, the RL policy generates corrective residuals $\mathbf{a}_t$ that adjust the robot's impedance and posture. This allows the system to maintain a stable support base even when the payload mass significantly alters the system's combined center-of-mass (CoM) dynamics.

2) Asymmetric teacher-student distillation:
to bridge the gap between simulation and reality for practical perception, we employ an asymmetric distillation framework:
\begin{itemize}
    \item Privileged teacher policy ($\pi_{T}$): during training, the teacher policy has access to privileged information $\mathbf{s}_{priv}$, including exact object mass $m_{obj}$, CoM offset $\Delta \text{CoM}$, and ground-truth external disturbances. The teacher policy establishes a high-performance baseline by directly mapping specific physical parameters to stabilizing joint actions.
    \item Adaptive student policy ($\pi_{S}$): the student policy relies solely on a proprioceptive history buffer $\mathbf{H}_t = \{ \mathbf{s}_{t-k}, \dots, \mathbf{s}_t \}$. Knowledge transfer is achieved by minimizing the KL divergence between policy distributions:
    \begin{equation}
        \min_{\phi} \mathbb{E}_{\mathbf{H}_t \sim \mathcal{D}} [ D_{KL} ( \pi_{T}(\cdot | \mathbf{s}_t, \mathbf{s}_{priv}) \| \pi_{S}(\cdot | \mathbf{H}_t; \phi) ) ]
    \end{equation}
\end{itemize}

The distillation process enables the student policy to map the patterns in joint-error histories and base vibrations to the corresponding compensatory actions learned by the teacher. This approach eliminates the need for explicit force-torque sensors or mass-estimation modules at runtime. By superposing the learned residuals onto the kinematically consistent RG prior, the IO-WBC framework maintains postural stability and motion synchronization across a wide range of payload conditions, ensuring that the system remains within its physical torque and friction limits.

\subsection{Interaction Robustness and Multi-Objective Reward}

We implement a rigorous randomization and reward scheme in a massively parallel simulation environment to ensure the stability of IO-WBC under non-stationary interactions.

1) Perturbation and domain randomization:
to simulate unpredictable physical impacts, we periodically inject random impulsive forces $\mathbf{I}_{rand}$ to the object's CoM. Furthermore, we apply domain randomization to object mass, friction coefficients $\mu$, and sensor latencies to enhance generalization across heterogeneous payloads.

2) Reward function design:
the total reward $\mathcal{R}$ is decomposed into four physical dimensions to guide the learning of safe and efficient transport:
\begin{itemize}
    \item Coupled motion consistency reward ($r_{sync}$):
    \begin{equation}
        r_{sync} =
        \alpha_1 e^{-\| \boldsymbol{\theta}_{\mathcal{S}} -
        \boldsymbol{\theta}_{\mathcal{S}}^{ref} \|^2}
        + \alpha_2 e^{-\| \mathbf{v}_{sys} - \mathbf{v}_{coll} \|^2} ,
    \end{equation}
    where $\mathbf{v}_{sys}$ denotes the translational velocity of the robot--object
    coupled system. This encourages the support branch to align with the RG prior and HRC commands.
    \item Object equilibrium reward ($r_{obj}$): penalizes 3D pose deviations of the object and vertical fluctuations relative to the base reference height to maintain stability.
    \item Whole-body smoothness reward ($r_{smooth}$): penalizes joint accelerations $\ddot{\mathbf{q}}$, action jitters $\|\mathbf{a}_t - \mathbf{a}_{t-1}\|^2$, and mechanical power consumption $\boldsymbol{\tau} \cdot \dot{\mathbf{q}}$.
    \item Gait constraint term ($r_{gait}$): provides positive guidance for foot swing height and contact timing based on the phase signal $\phi_t$.
\end{itemize}

3) Curriculum learning schedule:
we adopt a progressive curriculum that exposes the policy to increasingly strong robot–object coupling. Training progresses from static balance under varied loads, to dynamic locomotion with impulsive perturbations, and culminates in extreme commands where translational motion may be partially or fully infeasible.

\begin{figure*}[t]
    \centering
    \includegraphics[width=0.97\textwidth]{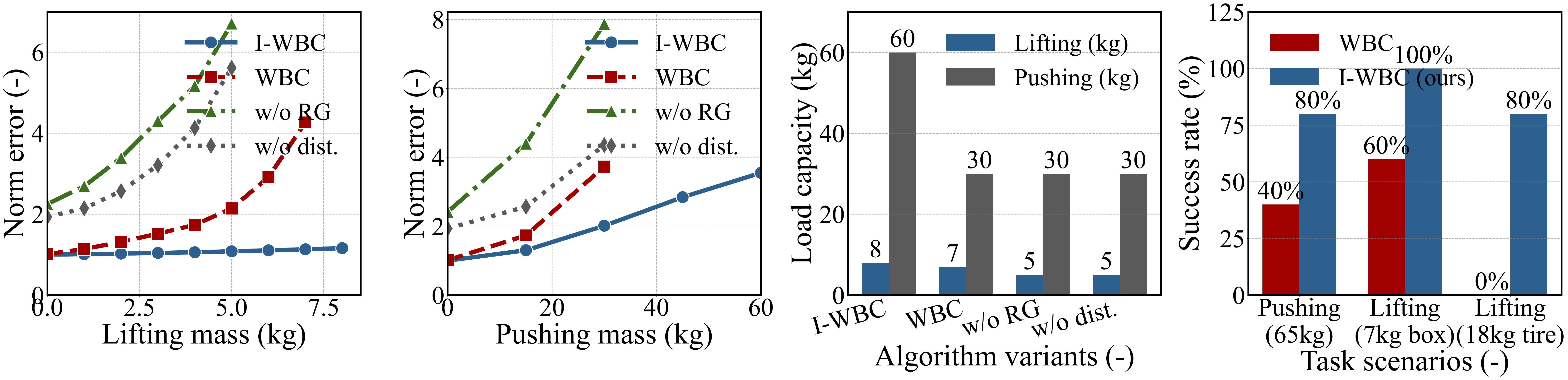}
    \caption{Comprehensive performance comparison across algorithmic variants. The first two panels report the norm errors in the lifting and pushing scenarios, while the two rightmost panels compare the load capacity and success rates.}
    \label{fig:four_comparison}
\end{figure*}

\section{Experimental Evaluation}
\label{sec:evaluation}

In this section, we evaluate the performance of the IO-WBC framework through extensive simulations and physical experiments. Our analysis focuses on how the lower-limb policy translates high-level HRC commands into robust locomotion and balanced postures under physical constraints.

\begin{table}[h]
\centering
\caption{Detailed hyperparameters for RL training and curriculum specifications.}
\label{tab:hyperparameters}
\small
\renewcommand{\arraystretch}{1.3} 
\begin{tabularx}{\columnwidth}{X c} 
\toprule
\textbf{Parameter} & \textbf{Value / Range} \\
\midrule
RG layers (hidden) & [128, 64] \\
IO-WBC policy layers (hidden) & [512, 256, 128] \\
Proprioceptive history ($H_t$) & 25 steps \\
Control frequency & 50 Hz \\
\midrule
Discount factor ($\gamma$) & 0.99 \\
GAE factor ($\lambda$) & 0.95 \\
Learning rate & $2 \times 10^{-4}$ \\
Number of parallel envs & 2048 \\
\midrule
Payload mass ($m_{obj}$) & $50\% \sim 150\%$ \\
Ground friction for robot ($\mu$) & $0.5 \sim 1.2$ \\
Ground friction for object ($\mu$) & $0.02 \sim 0.15$ \\
Surface friction for object ($\mu$) & $1.0 \sim 1.5$ \\
Target CoM height ($H_{\text{CoM}}$) & $0.5 \sim 0.8$ m \\
Torso pitch range ($\alpha_{\text{torso}}$) & $-0.5 \sim 1.5$ rad \\
Torso roll/yaw range & $-0.7 \sim 0.7$ rad \\
\bottomrule
\end{tabularx}
\end{table}

\begin{figure}[h]
    \centering
    \includegraphics[width=0.48\textwidth]{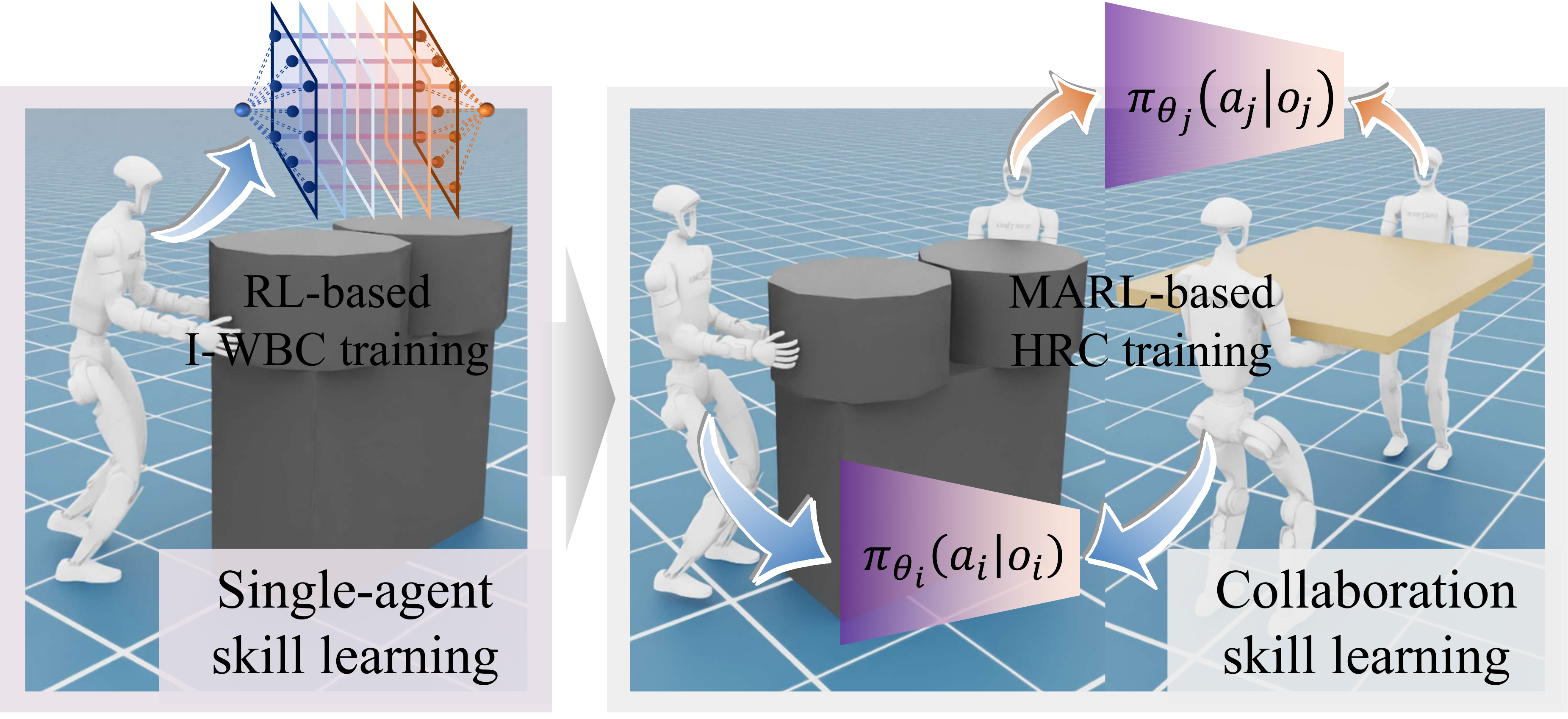}
    \caption{Visualization of the learning hierarchy in simulation. The framework encompasses single-agent skill learning for the IO-WBC layer and MARL for high-level HRC coordination.}
    \label{fig:sim_visualization}
\end{figure}

\begin{figure*}[t]
    \centering
    \includegraphics[width=0.999\textwidth]{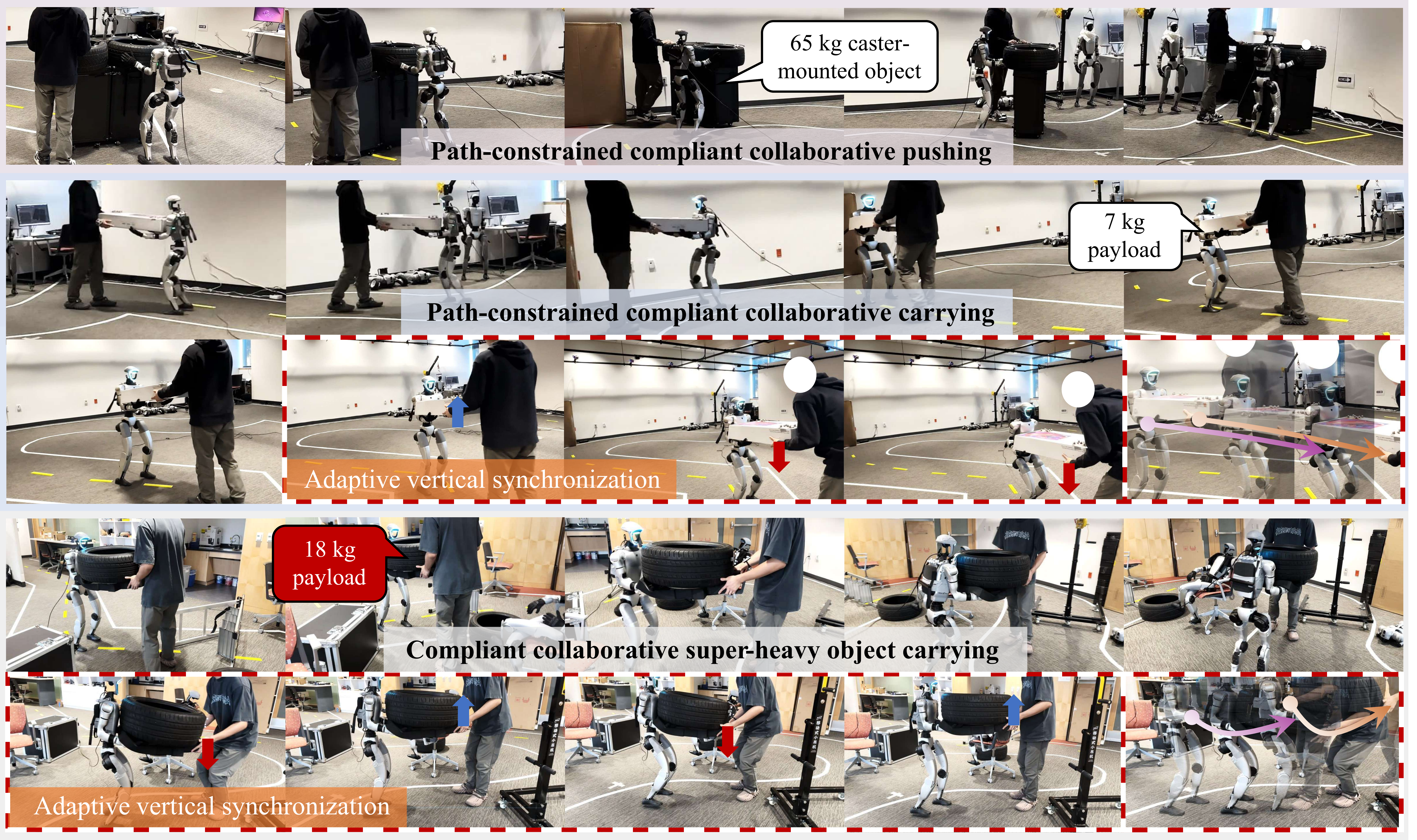}
    \caption{Real-world deployment of IO-WBC on the HRC tasks between human and Unitree G1, including path-constrained collaborative pushing and lifting, as well as collaborative super-heavy object lifting. Specifically, the red dashed regions highlight the adaptive vertical synchronization and postural resilience during the transport of a 7 kg box and an 18 kg tire.}
    \label{fig:real_world_results}
\end{figure*}

\begin{figure}[h]
    \centering
    \includegraphics[width=0.95\linewidth]{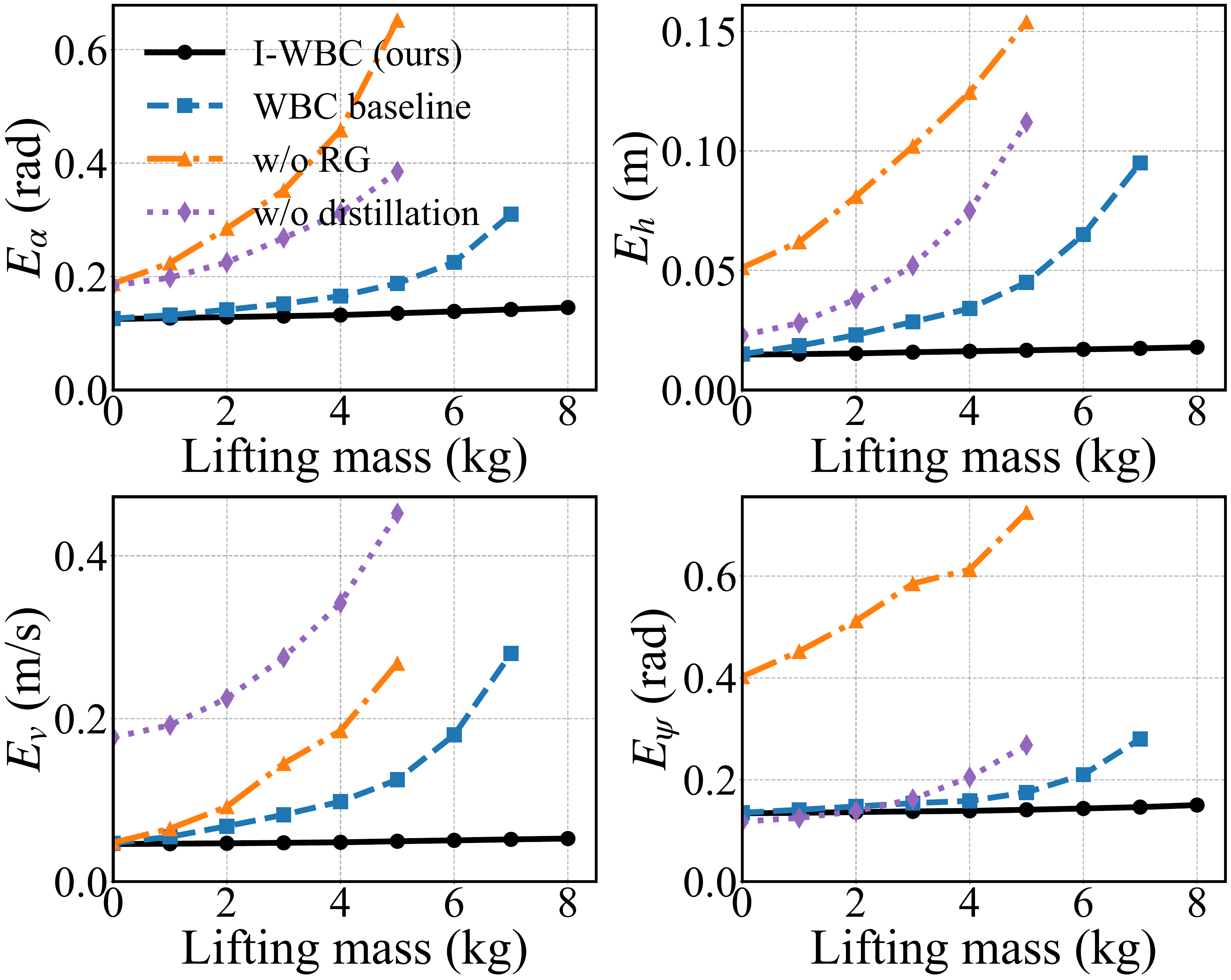}
    \caption{Tracking errors vs. lifting mass. IO-WBC maintains near-constant performance, while baselines exhibit exponential error growth in $E_{\alpha}, E_{h}, E_{v},$ and $E_{\psi}$.}
    \label{fig:lifting_errors}
\end{figure}

\subsection{Experimental Setup and Baselines}

The training is performed in the Isaac Lab \cite{mittal2023orbit}, and the physical experiments are conducted on a Unitree G1 robot cooperating with a human partner with reliance on a motion-capture (MoCap) system. The high-level tactical commands are generated by a multi-agent reinforcement learning-based skill policy, which treats the human-robot collaboration as a multi-agent Markov game. The detailed hyperparameters for the RL training and the hierarchical architecture are summarized in Table \ref{tab:hyperparameters}. The configuration of our domain randomization includes significant variances in link inertia, joint friction, and motor damping, following the mainstream protocols for humanoid whole-body control \cite{li2025amo} to ensure simulation-to-real transferability.

Task scenarios include compliant cooperative pushing and lifting:
1) Cooperative lifting: we test the limits of postural stability using two distinct payloads: a 7~kg cardboard box for nominal interaction and an 18~kg rubber tire for heavy-load-bearing trials. 
2) Cooperative pushing: the robot and a human partner transport a 65~kg crate across a smooth surface. To simulate low-friction, heavy-mass interaction, the crate is equipped with omni-directional wheels. This task evaluates the system's ability to maintain base stability while overcoming continuous non-conservative resistance.

Regarding the baselines for comparison, we evaluate IO-WBC (our framework) against three strong variants: 
1) WBC (SOTA): the full adaptive motion optimization (AMO-style) whole-body controller. Crucially, while this baseline represents a state-of-the-art model-based approach, its state observer and optimization kernels are inherently load-agnostic, assuming nominal robot dynamics without explicit coupling terms for heavy payloads.
2) w/o RG: a variant without the trajectory-optimized reference generator prior; 
3) w/o distillation: a variant trained without teacher--student asymmetric distillation. Importantly, the WBC baseline is not a simplified controller, but follows a high-performance adaptive motion optimization structure widely adopted in modern legged manipulation systems. It incorporates full-body kinematic consistency, constraint-aware posture tracking, and dynamic feasibility enforcement. The performance gap observed later thus stems specifically from the lack of interaction-awareness in classical optimization vs. our learned latent dynamics.

Performance is quantified using tracking error metrics $E_i$, defined as the mean absolute deviation between the reference command $r$ and the actual state $x$ over a task duration $T$:
\begin{equation}
    E_{i} = \frac{1}{T} \int_{0}^{T} |x_i(t) - r_i(t)| dt
\end{equation}
where the primary metrics evaluated are linear velocity error ($E_{v}$), base yaw error ($E_{\psi}$), torso pitch error ($E_{\alpha}$), and CoM height error ($E_{h}$).

\subsection{Performance Benchmark and Ablation Study}

To ensure the statistical reliability of the results, all real-world HRC trials and simulation benchmarks were conducted over 5 independent, successive runs. Performance metrics reported in this section represent the mean values, with success rates reflecting the aggregate performance across these trials to account for experimental variance. Ablation results, consolidated in the comparative analysis in Fig.~\ref{fig:four_comparison}, emphasize the synergy between the hierarchical layers. The w/o RG variant, lacking kinematic priors, exhibits $E_h$ and $E_{\psi}$ errors that are 2--4 times higher than the baseline even under nominal loads, as the policy must simultaneously solve for global balance and local interaction. Removing the distillation process (w/o Distillation) prevents the policy from decoding object dynamics, causing catastrophic collapse during heavy loading trials. This confirms that the latent representation of payload properties, captured through history-based distillation, is a prerequisite for stabilizing high-inertia interactions without explicit force sensing.

In real-world HRC trials, IO-WBC's resilience was quantified by its success rate. As shown in Fig.~\ref{fig:four_comparison}(c), IO-WBC achieved an 80\% success rate (4 out of 5 successful trials) in the critical 18 kg tire carrying task. In contrast, the WBC baseline consistently failed (0\% SR) due to its inability to proactively modulate lower-limb stiffness or shift its CoM projection to compensate for the 18 kg mass. This performance gap underscores IO-WBC's capacity for team-level synergy in assistive robotics tasks, where the robot effectively adapts to the human's loading intent.

\subsection{Postural Resilience in Lifting Tasks}

We conducted an incremental load-stress test from 0 kg to 8 kg with a 1 kg resolution across 5 independent trials per increment to observe the breakdown point of each strategy. As illustrated in Fig.~\ref{fig:lifting_errors}, IO-WBC exhibits superior linear stability across the entire gradient. While IO-WBC and WBC show near-identical performance at 0 kg, their trajectories diverge sharply as the mass increases beyond the nominal tuning range. Specifically, at the 4 kg tier, the baseline WBC's pitch error $E_{\alpha}$ begins to climb non-linearly. As the load approaches the 8kg critical limit, WBC and all other ablation variants suffer from divergent CoM oscillations in 100\% of the observed trials, leading to catastrophic failure. In contrast, IO-WBC maintains structural integrity with its pitch and height errors remaining nearly flat, demonstrating that implicit interaction-awareness effectively internalizes the payload's gravity as a learned prior. The low variance across the 5-trial set confirms that IO-WBC has learned a deterministic compensation strategy for high-inertia coupling. This stability is further reflected in the Norm error comparison shown in Fig.~\ref{fig:four_comparison}(a), where IO-WBC maintains a significantly lower error profile compared to the rapidly ascending curves of the baselines, as summarized in Table \ref{tab:lifting_huge}.

\begin{table}[h]
\centering
\caption{Detailed comparison of HRC command tracking errors under cooperative lifting (0 kg--8 kg).}
\label{tab:lifting_huge}
\small
\renewcommand{\arraystretch}{1.2}
\begin{tabularx}{\columnwidth}{l X c c c c}
\toprule
\textbf{Mass} & \textbf{Algorithm} & $\mathbf{E_{\alpha}}$ ($\downarrow$) & $\mathbf{E_{h}}$ ($\downarrow$) & $\mathbf{E_{v}}$ ($\downarrow$) & $\mathbf{E_{\psi}}$ ($\downarrow$) \\ 
\midrule

\multirow{4}{*}{0 kg} 
 & IO-WBC & 0.125 & 0.015 & 0.046 & 0.134 \\
 & WBC           & 0.126 & 0.015 & 0.047 & 0.135 \\
 & w/o RG       & 0.188 & 0.051 & 0.048 & 0.403 \\
 & w/o Distill. & 0.184 & 0.023 & 0.177 & 0.118 \\ 
\midrule

\multirow{4}{*}{4 kg} 
 & IO-WBC & 0.132 & 0.016 & 0.048 & 0.139 \\
 & WBC           & 0.166 & 0.034 & 0.098 & 0.158 \\
 & w/o RG       & 0.458 & 0.125 & 0.185 & 0.612 \\
 & w/o Distill. & 0.312 & 0.075 & 0.342 & 0.205 \\ 
\midrule

\multirow{4}{*}{8 kg} 
 & IO-WBC & 0.145 & 0.018 & 0.053 & 0.150 \\
 & WBC           & Fails & Fails & Fails & Fails \\
 & w/o RG       & Fails & Fails & Fails & Fails \\
 & w/o Distill. & Fails & Fails & Fails & Fails \\ 
\bottomrule
\end{tabularx}
\end{table}

\begin{table}[h]
\centering
\caption{Detailed comparison of HRC command tracking errors under cooperative pushing (0 kg--60 kg).}
\label{tab:pushing_huge}
\small
\renewcommand{\arraystretch}{1.2} 
\begin{tabularx}{\columnwidth}{l X c c c c}
\toprule
\textbf{Mass} & \textbf{Algorithm} & $\mathbf{E_{\alpha}}$ ($\downarrow$) & $\mathbf{E_{h}}$ ($\downarrow$) & $\mathbf{E_{v}}$ ($\downarrow$) & $\mathbf{E_{\psi}}$ ($\downarrow$) \\ 
\midrule

\multirow{4}{*}{0 kg} 
 & IO-WBC & 0.125 & 0.015 & 0.046 & 0.134 \\
 & WBC           & 0.126 & 0.015 & 0.047 & 0.135 \\
 & w/o RG       & 0.286 & 0.049 & 0.048 & 0.413 \\
 & w/o Distill. & 0.184 & 0.023 & 0.177 & 0.118 \\ 
\midrule

\multirow{4}{*}{30 kg} 
 & IO-WBC & 0.131 & 0.016 & 0.225 & 0.143 \\
 & WBC           & 0.211 & 0.055 & 0.356 & 0.246 \\
 & w/o RG       & 0.582 & 0.165 & 0.453 & 0.785 \\
 & w/o Distill. & 0.323 & 0.073 & 0.386 & 0.211 \\ 
\midrule

\multirow{4}{*}{60 kg} 
 & IO-WBC & 0.152 & 0.019 & 0.486 & 0.161 \\
 & WBC           & Fails & Fails & Fails & Fails \\
 & w/o RG       & Fails & Fails & Fails & Fails \\
 & w/o Distill. & Fails & Fails & Fails & Fails \\ 
\bottomrule
\end{tabularx}
\end{table}

\subsection{Effectiveness and Compliance in Heavy-Duty Pushing}

\begin{figure}[h]
    \centering
    \includegraphics[width=0.95\linewidth]{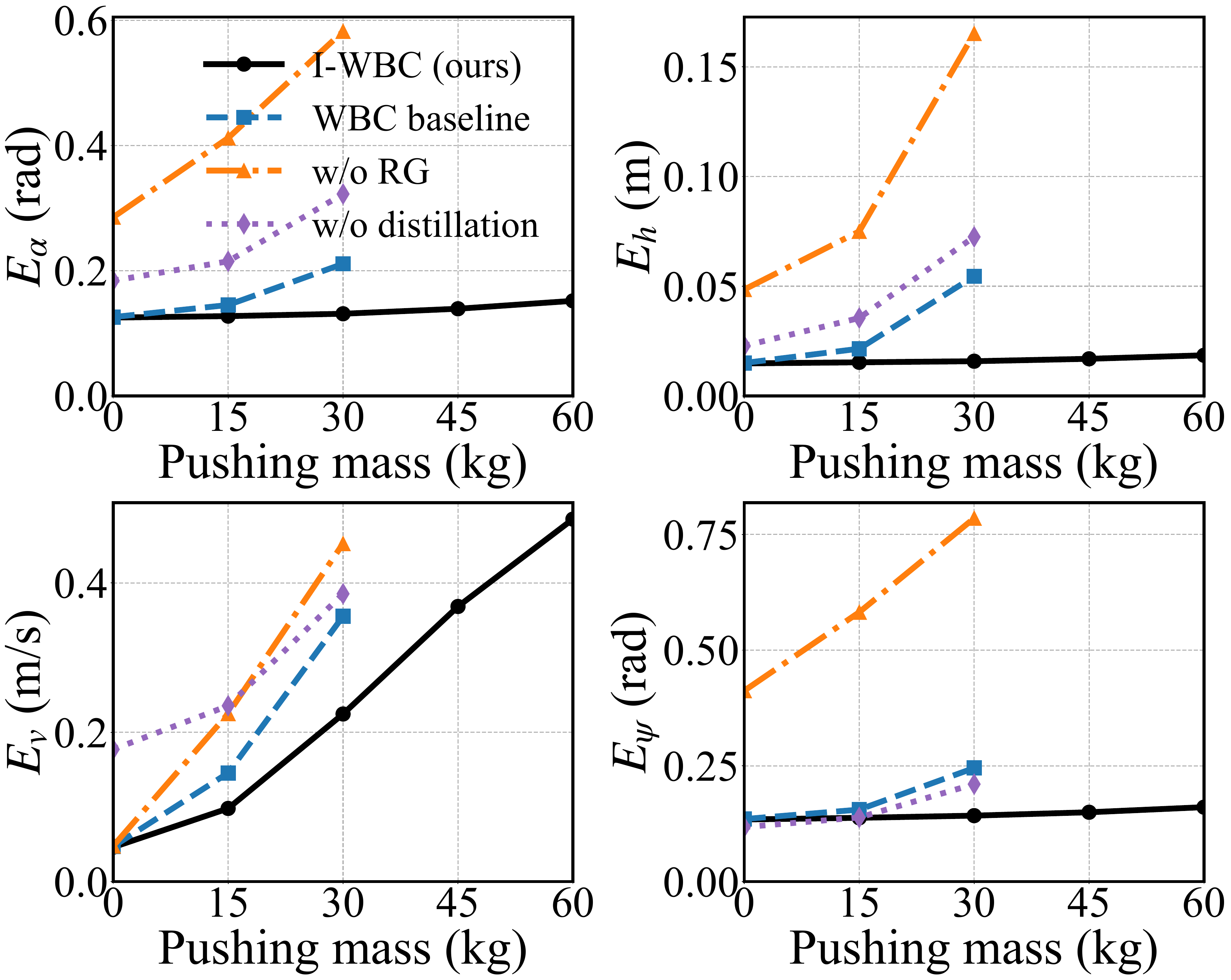}
    \caption{Tracking errors vs. pushing mass. Despite rising velocity errors ($E_{v}$), IO-WBC preserves high-fidelity postural control ($E_{\alpha}, E_{h}$) up to 60 kg.}
    \label{fig:pushing_errors}
\end{figure}

The 60 kg pushing task highlights the trade-off between command following and physical survival. As shown in Fig.~\ref{fig:pushing_errors}, under extreme resistance, the robot's locomotion enters a compliant-stagnation regime. As the load increases to 60 kg, the velocity error $E_v$ for IO-WBC grows significantly, reaching approximately 0.486 m/s. This increase in $E_v$ is not an algorithmic failure, but a strategic prioritization of postural invariance ($E_{\alpha}, E_{h}$) over locomotion speed. This stagnation reflects a policy-level trade-off between velocity tracking and support stability under extreme resistance. Rather than saturating joint torques or inducing foot slippage—which caused the WBC baseline to fail in all 5 trials for 60 kg level—the learned controller reduces forward progression to preserve the robot's center-of-mass within the friction cone. In contrast, IO-WBC survives up to 60 kg by actively modulating its impedance. As evidenced in Fig.~\ref{fig:four_comparison}, while baseline errors explode beyond 30kg, the IO-WBC error norm grows linearly and remains bounded, proving its robustness at the edge of physical actuator limits, as summarized in Table \ref{tab:pushing_huge}.

\section{Conclusion}

This paper presented an interaction-oriented whole-body control framework that enables humanoid robots to maintain stable force application and postural integrity during heavy-load human--robot collaborative transport. By regulating whole-body posture and residual interaction compensation, the proposed approach ensures physically consistent execution of skill-level commands even under extreme robot--object coupling where analytical models typically deviate.

\begin{itemize}
\item We introduced a hierarchical interaction-oriented WBC that synergizes a trajectory-optimized reference generator with an interaction-aware residual policy. This architecture ensures consistent whole-body stabilization by bridging the gap between kinematic feasibility and dynamic uncertainty, enabling the robust execution of skill-level commands under high-inertia coupling.

\item A physics-aware learning approach with asymmetric teacher--student distillation is developed, facilitating the implicit identification of interaction dynamics from proprioception alone. This allows the policy to maintain stable behavior under increasing payloads where state-of-the-art WBC strategies become physically infeasible.

\item Extensive experiments across 5 independent trials per task validated significant robustness improvements. IO-WBC achieved 80\% success in transporting an 18\,kg payload while the baseline failed, maintained support robustness up to 8\,kg lifting where all baselines collapsed, and remained operational under pushing loads up to 60\,kg with consistent postural invariance.
\end{itemize}








\bibliographystyle{IEEEtran}
\bibliography{IEEEexample}



\end{document}